\documentclass[letterpaper, 10 pt, journal, twoside]{IEEEtran}                                   

\usepackage{graphics} 
\usepackage{epsfig} 
\usepackage{mathptmx} 
\usepackage{times} 
\usepackage{amsmath} 
\usepackage{amssymb}  
\usepackage{algorithm}
\usepackage{algorithmicx}
\usepackage[]{algpseudocode}
\usepackage{cite}
\usepackage{subcaption}
\usepackage{booktabs}
\usepackage{rotating}
\usepackage{multirow}
\usepackage{array}
\usepackage[table]{xcolor}
\usepackage{balance}
\usepackage{comment}
\usepackage{paralist}
\usepackage{todonotes}
\usepackage{microtype}
\usepackage{dblfloatfix}
\usepackage{hyperref}

\usepackage{soul}

\begin{document}

\title{\LARGE \bf Estimating Trust in Human-Robot Collaboration through\\Behavioral Indicators and Explainability}

\author{Giulio Campagna$^{1}$, Marta Lagomarsino$^{2}$, Marta Lorenzini$^{2}$, Dimitrios Chrysostomou$^{3}$, \\Matthias Rehm$^{1}$, Arash Ajoudani$^{2}$ \vspace{-0.5cm}
\thanks{Manuscript received: February, 18, 2025; Revised May, 30, 2025; Accepted August, 4, 2025. This paper was recommended for publication by Editor Angelika Peer upon evaluation of the Associate Editor and Reviewers' comments. The work was supported in part by the European Union Horizon Project TORNADO (Grant No. 101189557) and in part by the Independent Research Fund Denmark (Grant No. 1032-00311B).}
\thanks{This work involved human subjects in its research. Approval of all ethical and experimental procedures and protocols was granted by ASL3 Genovese under Application No. IIT\_HRII\_ERGOLEAN 156/2020, and performed in line with the Helsinki Declaration.}
\thanks{Corresponding author's email: {\tt\footnotesize gica@create.aau.dk}}
\thanks{$^{1}$ 
Human-Robot Interaction Lab., Technical Faculty of IT and Design, Aalborg University, Denmark.}
\thanks{$^{2}$  
Human-Robot Interfaces and Interaction Lab., Istituto Italiano di Tecnologia (IIT), Italy.}
\thanks{$^{3}$ 
Smart Production Lab., Faculty of Engineering and Natural Sciences, Aalborg University, Denmark.}
\thanks{Digital Object Identifier (DOI): see top of this page.}
}

\markboth{IEEE Robotics and Automation Letters. Preprint Version. Accepted August, 2025}
{Campagna \MakeLowercase{\textit{et al.}}: Estimating Trust via Behavioral Indicators and Explainability}

\maketitle
\IEEEpeerreviewmaketitle

\begin{abstract}

Industry 5.0 focuses on human-centric collaboration between humans and robots, prioritizing safety, comfort, and trust. This study introduces a data-driven framework to assess trust using behavioral indicators. The framework employs a Preference-Based Optimization algorithm to generate trust-enhancing trajectories based on operator feedback. This feedback serves as ground truth for training machine learning models to predict trust levels from behavioral indicators. The framework was tested in a chemical industry scenario where a robot assisted a human operator in mixing chemicals. Machine learning models classified trust with over 80\% accuracy, with the Voting Classifier achieving 84.07\% accuracy and an AUC-ROC score of 0.90. These findings underscore the effectiveness of data-driven methods in assessing trust within human-robot collaboration, emphasizing the valuable role behavioral indicators play in predicting the dynamics of human trust.

\end{abstract}


\section{INTRODUCTION}
\label{sec:introduction}

Industry 5.0 shifts to human-centered manufacturing, where collaborative robots work alongside human workers to enhance productivity and efficiency. By integrating technologies like sensors and machine learning, robots adapt to human actions and dynamic environments. Focusing on safety, adaptability, and Human-Robot Collaboration (HRC), it creates a resilient, sustainable industrial ecosystem that combines human decision-making with robotic precision for improved performance~\cite{gervasi2023experimental}. Trust is key for ensuring safe, comfortable, and seamless human-robot interaction.

Lee and See~\cite{lee2004trust} define trust as the belief that an agent will support a person’s goals, particularly in uncertain or vulnerable situations. Confidence in a robot’s ability to perform tasks reliably is critical to effective human-robot interaction. Under-trust may lead to operator overload by undervaluing the robot’s capabilities, while over-trust can cause safety risks such as equipment damage or collisions~\cite{de2020towards}.
Trust is often assessed via post-interaction surveys (e.g.,~\cite{schaefer2016measuring}), which fail to capture real-time fluctuations and may not reflect actual user behavior~\cite{campagna2025systematic}. This highlights the need for online trust estimation to enable adaptive robot behavior that ensures safety and supports ergonomic collaboration.

Recently, data-driven approaches have emerged as effective solutions for online trust estimation. Xu and Dudek~\cite{xu2015optimo} introduced a dynamic Bayesian network to continuously estimate human trust in robots based on task performance and factors like failure rates, human interventions, and task outcomes. Shayesteh et al.\cite{shayesteh2022workers} developed a method to evaluate trust in construction robots during collaborative tasks using EEG signals as input to machine learning models, proving highly effective. In our recent studies, body motion data~\cite{campagna2024data} and facial expressions~\cite{campagna2024analysis} were independently utilized as inputs to machine learning models for estimating trust levels in a chemical industry scenario, where a robot assisted a human operator with chemical delivery and mixing tasks. 
Finally, Lagomarsino et al.~\cite{lagomarsino2023maximising} proposed a reinforcement learning framework that adjusts interaction parameters based on a human-robot coefficiency metric, integrating human physical and cognitive factors, and robot operational costs to optimize joint efficiency, align with user preferences, enhance comfort, and foster trust.

Despite recent advancements, data-driven trust estimation models remain underexplored and struggle to capture the full complexity of trust dynamics~\cite{campagna2025systematic}. Many rely on single-factor indicators, overlooking the integration of behavioral, cognitive, and contextual features that are critical for accurate modeling. Moreover, most existing models lack personalization, failing to account for individual differences—such as personality traits, emotional responses, or past experiences—which are essential for adapting robot behavior to the unique trust-building needs of each user. This limits the potential for fostering long-term, meaningful interactions. In addition, many models are not seamlessly integrated into robotic systems, preventing online behavioral adaptation in response to fluctuating trust levels. Leveraging behavioral trust indicators offers a promising path forward by enabling deeper insight into trust dynamics and supporting more adaptive, context-aware collaboration.

\begin{figure}
    \vspace{0.1cm}
    \centering
    \includegraphics[width=0.65\linewidth]{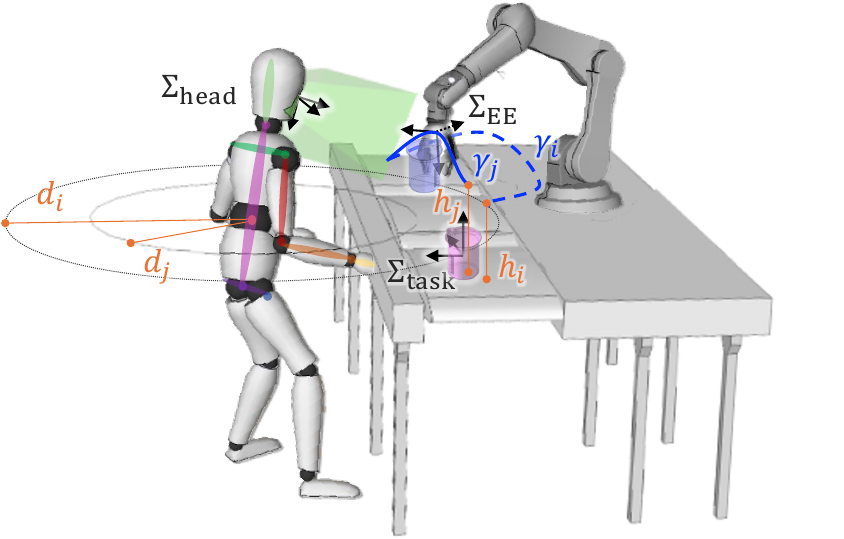}
    \caption{
    Illustration of HRC in an industrial setting, highlighting operator whole-body tracking (head and upper body) and interaction parameters guiding the robotic manipulator's behavior.
  \vspace{-0.4cm} }
    \label{framework_parameters}
\end{figure}

This work introduces a data-driven framework that extends beyond our previous study~\cite{campagna2024promoting} by integrating human- and robot-related behavioral trust indicators with machine learning models to estimate human trust preferences online. While the \textit{Preference-Based Optimization} (PBO) algorithm from our prior work is employed solely to generate robot trajectories based on explicit human feedback, the key innovation here lies in leveraging these behavioral indicators to train a machine learning model. 
The model predicts trust preferences by analyzing the dynamic relationship between interaction parameters and trust indicators, enabling a more personalized and adaptive approach to HRC. Trust indicators are derived from human motion capture and robot motion characteristics during collaboration. To enhance interpretability, SHAP (SHapley Additive exPlanations) values are used to explain model predictions, highlighting how each indicator influences overall trust estimation and how these effects vary across individual human operators.
The framework's effectiveness is evaluated in a chemical industry scenario, where a robotic arm assists a human in performing the potentially hazardous task of pouring chemicals.

The main contribution of this work is twofold: 

\begin{itemize}
    \item The development of a novel data-driven model that leverages both human and robot behavior to predict human trust preferences, enabling the optimization of key interaction parameters—specifically, the robot’s execution time, the separation distance between human and robot, and the vertical proximity of the end-effector to the user’s head—during collaborative tasks.
    \item A comprehensive analysis of behavioral trust indicators, with a focus on explainability and model-driven insights into which indicators most influence trust and how their effects differ across individuals.
\end{itemize}

The remainder of this paper is structured as follows: Section~\ref{sec:methodology} details the methodology, followed by the experimental procedure in Section~\ref{sec:experiments}. Section~\ref{sec:results} presents the results, and 
Section~\ref{sec:discussion} provides the discussion and conclusions of the work.

\section{METHODOLOGY}
\label{sec:methodology}

This section defines behavioral indicators from human movement sensor data during human-robot interactions to study human trust dynamics in robots. We then describe a machine-learning pipeline that uses these indicators to estimate human trust preferences for robot trajectories. Using PBO, we collected explicit user feedback comparing two interaction parameter sets shaping robot trajectories, providing ground-truth labels for model training. Finally, we evaluate model explainability to reveal how each behavioral indicator influences trust, identifying key factors shaping user preferences and highlighting individual variations. Understanding these trends enables adaptive, personalized robot behavior to foster trust effectively in human-robot interactions.

This paper proposes a rigorous pipeline to address the following research questions (RQs):

\begin{enumerate}[\bf RQ1.]
    \item \label{RQ1} \textit{Can behavioral indicators and machine learning models predict user preferences for robot interaction parameters that promote trustworthy HRC?}
    \item \label{RQ2} 
    \textit{What is the contribution of each indicator to the estimation of trust levels?} 
\end{enumerate}

\subsection{Definition of the Trust Indicators}
\label{sec:indicators_definition}

Several indicators are proposed to capture trust dynamics in industrial HRC settings. These indicators are categorized into \textit{human-related factors} and \textit{robot-related factors}, following the categorization scheme used in~\cite{hancock2011meta}. By combining these parameters, the framework aims to provide a holistic view of trust, analyzing both human body language and perception of robot behaviors. 
The framework operates without assuming the relevance or direction of correlations between behavioral indicators and perceived trust. Instead, these relationships are determined by the machine learning model and explained through detailed analysis, forming the foundation for online trust modeling and effective collaboration.

\subsubsection{Human-Related Trust Indicators}
\label{sec:human_indicators}

Human-related trust indicators reflect the cognitive and behavioral responses of the human operator during robot interaction, offering insights into their trust and comfort levels. The human body is modeled as a kinematic chain with \( N \) joints and \( N + 1 \) rigid body segments, each associated with a frame \( \Sigma_i \), where \( i \in \{1, \dots, N\} \), relative to a global reference frame \( \Sigma_W \). The global frame's position is calibrated initially using the motion tracking system, ensuring accurate tracking over time.

\textbf{Human Attention to End-Effector} -  The proposed trust indicator examines the attention an individual directs toward the robot during a collaborative task. 
Insights from contemporary psychology suggest that average dwell time—defined as the mean duration a person’s gaze remains fixed on a specific area, such as the robot’s end-effector—decreases with increasing confidence and expertise \cite{holmqvist2011eye}. Based on this premise, attention is a relevant parameter for assessing trust. 

To evaluate the level of attention toward the end effector, we adopt the method proposed in~\cite{lagomarsino2022pick}. A frame, namely \( \Sigma_{\text{head}} \), is positioned at the center of the head and tilted by ten degrees to approximate the direction of the gaze~\cite{weidenbacher2007comprehensive}. The Cartesian vector representing the relative position between \( \Sigma_{\text{gaze}} \) (the estimated gaze frame) and \( \Sigma_{\text{EE}} \) (the end-effector frame) is converted into spherical coordinates, characterized by the azimuth angle \( \theta_{EE} \in \mathbb{R}\), the elevation angle \( \phi_{EE} \in \mathbb{R}\), and the radial distance $d_\text{EE} \in \mathbb{R}$. 
A fuzzy logic function that utilizes a Raised-Cosine Filter is then applied separately to normalize the attention angles $\theta_{EE}$ and $\phi_{EE}$, which are denoted here as $\alpha(t)$:

\vspace{-0.4cm}
{\small
\begin{equation}
\lambda(\alpha(t)) =
\begin{cases} 
1, & \text{if } |\alpha(t)| \leq \alpha_{\text{min}}(t), \\[6pt]
\dfrac{1}{2} \left[ 1 - \cos\left( 
    \frac{|\alpha(t)| - \alpha_{\text{min}}(t)}
         {\alpha_{\text{max}}(t) - \alpha_{\text{min}}(t)} \pi 
\right) \right], 
& \text{if } |\alpha(t)| > \alpha_{\text{min}}(t) \\ 
  & \text{\& } |\alpha(t)| \leq \alpha_{\text{max}}(t), \\[6pt]
0, & \text{otherwise}.
\end{cases}
\label{fuzzy_logic_EE}
\end{equation}
}

The threshold values \( \alpha_{\text{min}}(t) \) and \( \alpha_{\text{max}}(t) \) depend on the current distance \( d_\text{EE} \) of the end-effector from the human operator. These thresholds are calculated using the approach described in~\cite{lagomarsino2024promind}:

\vspace{-0.3cm}
{\small
\begin{equation}
\alpha_{\text{min}}(t) = \tan^{-1} \left( \frac{(1 - \delta) r_\text{EE} }{d_\text{EE}(t)} \right), \quad
\alpha_{\text{max}}(t) = \tan^{-1} \left( \frac{(1 + \delta) r_\text{EE}}{d_\text{EE}(t)} \right),
\label{thresholds_alpha_EE}
\end{equation}
}

\noindent where \( r_\text{EE} \in \mathbb{R} \) represents a fixed predefined radius of the end-effector's containment area. 
The parameter \( \delta \in \mathbb{R} \) is set to \(  0.4 \) to ensure smooth variation in the function \( \lambda(\alpha(t)) \).

In conclusion, the attention level \( \Lambda_\text{EE}(t) \in [0,1] \) directed toward the end-effector is defined as the product of the normalized azimuth and elevation indicators:

\vspace{-0.4cm}
\begin{equation}
\Lambda_\text{EE}(t) = \lambda(\theta_\text{EE}(t)) \lambda(\phi_\text{EE}(t)).
\label{attention}
\end{equation}

When \( \Lambda_{EE} \approx 1 \), the human is actively monitoring the robot's motion; when \( \Lambda_{EE} \approx 0 \), the human shows no focus on the robot.

\textbf{Human Attention to Task} - This trust indicator monitors the level of attention the human directs toward the task itself.
Similar to the \textit{Human Attention to End-Effector} indicator, we compute the relative position between \( \Sigma_{\text{gaze}} \) and a fixed reference frame \( \Sigma_{\text{task}} \), which defines the area where the task is performed and uses $r_\text{task}$ to delimit the task area. Consequently, we apply Eq. \eqref{fuzzy_logic_EE} and \eqref{attention} to evaluate the current level of attention $\Lambda_\text{task}(t) \in [0,1]$. 

\textbf{Spatial Displacement} - This indicator evaluates the human operator's spatial deviation from the initial position. 
The spatial dynamics of human-robot interaction offer valuable insights into trust. Some studies suggest that increased human displacement may indicate discomfort, stress, or uncertainty about the robot’s performance~\cite{campagna2023analysis,campagna2024data}. Conversely, other research shows that a lack of confidence in the robot can cause humans to "freeze", waiting for the robot to finish its movement before continuing the task~\cite{sweller2011cognitive}.

To calculate this indicator, the displacement \( \Delta p \in \mathbb{R} \) is defined as the normalized distance between the current position of the human head, \( \mathbf{p}_{\text{head}} \in \mathbb{R}^3 \), and its initial position at the start of the task, \( \mathbf{p}_{\text{init}} \in \mathbb{R}^3 \): 

\vspace{-0.2cm}
\begin{equation}
\Delta p(t) = \frac{\|\mathbf{p}_{\text{head}}(t) - \mathbf{p}_{\text{init}}\|}{d_{\text{max}}},
\label{displacement_comfort}
\end{equation}

\noindent where \( d_{\text{max}} \in \mathbb{R} \) is the maximum allowable distance between the head and the fixed point. This ensures that \( \Delta p(t) \) is normalized to the range \( [0, 1] \), with \( \Delta p(t) \approx 0 \) corresponding to minimal head movement and \( \Delta p(t) \approx 1 \) indicating significant motion.

\textbf{Human-Robot Speed Synchronization} - Previous studies have shown that smooth coordination and synchronized movements between humans and robots are strongly correlated with a positive perception and trust in the robot. Conversely, significant speed discrepancies indicate misalignment and potential trust issues~\cite{kanda2003Body, bartkowski2023sync}. To address this, we define an indicator that measures the alignment between the speed of the human hand and the robot's end-effector during the task. 

The relative speed match, \( \Delta \text{v} \in [0,1]\), is computed as:

\vspace{-0.2cm}
\begin{equation}
\sigma(t) = 1 - \left| \frac{ \text{v}_{\text{hand}}(t) - \text{v}_{\text{min},\text{hand}} }{\text{v}_{\text{max},\text{hand}} - \text{v}_{\text{min},\text{hand}}} - \frac{\text{v}_{\text{EE}}(t) - \text{v}_{\text{min},\text{EE}} }{\text{v}_{\text{max},\text{EE}} - \text{v}_{\text{min},\text{EE}}} \right|,
\label{hand_synchro}
\end{equation}

\noindent where \( v_{\text{hand}} \in \mathbb{R} \) and \( v_{\text{EE}} \in \mathbb{R} \) represent the magnitudes of the human hand velocity and the robot end-effector velocity, respectively. 
\( v_{\min,\text{hand}} \) and \( v_{\max,\text{hand}} \) denote the minimum and maximum observed velocity values for the human hand, while \( v_{\min,\text{EE}} \) and \( v_{\max,\text{EE}} \) represent the lower and upper bounds of the robot end-effector velocity. As a result, \( \sigma \approx 0 \) indicates a significant speed mismatch, while larger values signify closely matched velocities. 

\textbf{Reaction Time} - This measure quantifies the temporal latency between the robot's actions and the human operator's responses. Research suggests that tasks involving complex decision-making may increase delays in movement initiation~\cite{khan2006inferring}, reflecting heightened cognitive demands. As a key trust factor, longer reaction times may denote uncertainty or reduced confidence in the robot's performance.

Let \( t_R \in \mathbb{R} \) represent the timing of the robot's motion initiation, and \( t_H \in \mathbb{R} \) the human's action start time. 
The temporal misalignment \( \Delta t \in [0,1] \) is defined as:

\vspace{-0.2cm}
\begin{equation}
\Delta t = \frac{|t_H - t_R|}{\tau},
\label{reaction_time}
\end{equation}

\noindent where $\tau \in \mathbb{R}$ is the total execution time of the robot's trajectory.
A value of \( \Delta t \approx 0 \) indicates near-perfect synchronization, while \( \Delta t \approx 1 \) suggests a significant delay in the human's response to the robot.

\subsubsection{Robot-Related Trust Indicators}
\label{sec:robot_indicators}

Robot-related trust indicators are calculated based on the robot's behavioral characteristics, which are known to influence human perception and trust in the interaction. 

\textbf{Predictability} - 
Existing research has demonstrated that smooth robot maneuvers with low jerk values enhance predictability by aligning with human expectations, thereby increasing operator confidence and fluidity in interaction~\cite{kuehnlenz2016reduction, lagomarsino2022robot}. This effect is likely due to their similarity to natural arm motions~\cite{flash1985coordination}. In contrast, high-jerk movements can appear erratic, reducing predictability and trust. 

To quantify the impact of robot smoothness on human trust, an indicator is defined as the normalized integral of the squared jerk over the time interval \([t_0, t_f]\), where \( t_0 \) and \( t_f \) represent the starting and ending instants of time of the robot's trajectory. The jerk \(j(t) \in \mathbb{R}\), which is the magnitude of the third derivative of position, is constrained by the feasible minimum and maximum jerk values of the robot's end effector, \( j_{\text{min}}\) and \( j_{\text{max}} \in \mathbb{R}\). The \textit{Predictability} indicator \( \rho(t) \in [0,1] \) is then computed as:

\vspace{-0.1cm}
\begin{equation}
\rho(t) = 1 - \frac{\int_{t_0}^{t_f} j(t)^2 dt -  j_{\text{min}}^2 \tau_{\text{min}} }{ j_{\text{max}}^2 \tau_{\text{max}} -  j_{\text{min}}^2 \tau_{\text{min}} },
\label{predictability}
\end{equation}

\noindent where \( \tau_{\text{min}} \) and \( \tau_{\text{max}} \) represent the minimum and maximum durations of the robot's trajectory.
A \( \rho \approx 0 \) value reflects significant jerk fluctuations, while \( \rho \approx 1 \) signifies smooth and consistent motion.

\textbf{Legibility} - 
This factor evaluates the curvature of the robot's trajectory, which is used in literature as a measure of legibility—how clearly the robot's motion communicates its intent or goal to a human observer based on a partially observed trajectory~\cite{
dragan2013legibility}.

The curvature coefficient \( \kappa \in \mathbb{R} \) of the robot end-effector trajectory \( \gamma(t) \) is computed as:

\begin{equation}
\kappa(t) = \frac{\sqrt{\|\dot{\gamma}(t)\|^2 \|\ddot{\gamma}(t)\|^2 - (\dot{\gamma}(t) \cdot \ddot{\gamma}(t))^2}}{\|\dot{\gamma}(t)\|^3},
\label{curvature_coefficient}
\end{equation}

\noindent where \( \dot{\gamma}(t) \) and \( \ddot{\gamma}(t) \) are the first and second derivative, representing the robot's velocity and acceleration, respectively. 
The radius of curvature \( r \in \mathbb{R} \), which reflects the size of the approximating circle at a point on the trajectory, is then calculated as \( \left( \frac{1}{k} \right) \). Using the vector cross product identity \( \left| \mathbf{a} \times \mathbf{b} \right|^2 = \left| \mathbf{a} \right|^2 \left| \mathbf{b} \right|^2 - (\mathbf{a} \cdot \mathbf{b})^2 \), we obtain:

\vspace{-0.2cm}
\begin{equation}
r(t) = \frac{\|\dot{\gamma}(t)\|^3}{\|\dot{\gamma}(t) \times \ddot{\gamma}(t)\|}.
\label{radius_curvature}
\end{equation}

The \textit{Legibility} indicator \( L \in [0, 1] \) is obtained by scaling the radius relative to its maximum value \( r_{\text{max}} \):
\begin{equation}
L(t) = \frac{r(t)}{r_{\text{max}}}.
\label{legibility}
\end{equation}

\noindent A value of \( L \approx 0 \) corresponds to a robot trajectory with a smaller radius of curvature.
In contrast, when \( L \approx 1 \), the trajectory is associated with a larger radius of curvature, potentially making its actions easier to interpret.

\subsection{Dataset Collection}
\label{sec:data_collection}

To evaluate the relevance of the behavioral indicators for modeling human trust in the robot and confidence in HRC, the study relies on reliable ground truth data. As obtaining absolute trust values in realistic scenarios is challenging, the adaptive trajectory framework presented in~\cite{campagna2024promoting} is adopted. This framework employs the PBO algorithm~\cite{bemporad2021global} to iteratively refine the interaction parameters that shape the robot’s trajectory, based on repeated explicit trust feedback. At each iteration, new parameters are selected by minimizing an acquisition function that balances surrogate model exploitation and exploration, as detailed in~\cite{campagna2024promoting}.

\subsubsection{Preference Labels}
Specifically, for a robot's trajectory from point \textbf{A} to point \textbf{B}, three key interaction parameters (refer to Fig.~\ref{framework_parameters}) that influence  the user's trust are adjusted to align the robot's behavior with human preferences and expectations:

\begin{enumerate}[i)]
    \item the \textit{total execution time} \(\tau\) of the trajectory, which determines the robot's velocity profile;
    \item the \textit{separation distance} \(d\) maintained by the robot from the user's body;
    \item the \textit{maximum height} \(h\), controlling the vertical proximity of the robot's end effector to the user’s head.
\end{enumerate}

\noindent These parameters form the decision vector \(\mathbf{x} = [\tau, d, h] \in \mathbb{R}^3\). At the first iteration, the PBO algorithm proposes two distinct parameter sets \( \mathbf{x}_1, \mathbf{x}_2 \in X \), where \( \mathbf{x}_1 \neq \mathbf{x}_2 \). The human operator is then asked to indicate a preference using the \textit{preference function} $\pi: \mathbb{R}^n \times \mathbb{R}^n \to \{-1, 1\} $, defined as:

\vspace{-0.2cm}
\begin{equation}
    \pi(\textbf{x}_1, \textbf{x}_2) =
    \begin{cases}
        -1  & \text{if } \textbf{x}_1 \text{ is preferred over } \textbf{x}_2 \\
        1 & \text{if } \textbf{x}_2 \text{ is preferred over } \textbf{x}_1
    \end{cases}
    \label{pref_function}
\end{equation}

Each parameter set $\textbf{x}_i$ refines the robot's trajectory $\gamma$ to reach point \textbf{B} as follows: i) the total trajectory duration is set to \(\tau_i\); ii) a circle with radius \(d_i\) around the human defines a protected zone, with waypoints inside this zone adjusted to lie on the circle's edge; iii) the \(z\)-component of the trajectory is modified so that the motion's maximum height corresponds to \(h_i\). Details on trajectory adaptation can be found in~\cite{campagna2024promoting}.

In each iteration, a new trajectory $\gamma$, defined by a parameter set $\textbf{x}_i$, is proposed to the human operator. The operator selects the trajectory that best promotes trust, either by retaining the previous best trajectory (assigning $-1$ to $\pi$) or by adopting the refined trajectory (assigning $1$ to $\pi$). 
This iterative process identifies the optimal parameter vector, $\textbf{x}^*$, that minimizes the preference function $\pi$ within the bounded domain $[\mathbf{x}_{\text{min}}, \mathbf{x}_{\text{max}}]$, thereby maximizing trust in the robot.
This work focuses on participant preferences when comparing two robot behaviors. While these do not directly measure absolute trust, they provide valuable insight into trust dynamics across behaviors. Combined with human behavioral indicators, the labels indicate whether parameter changes increased or decreased trust and confidence.

\subsubsection{Data Pre-Processing and Trust Indicators Selection}
\label{sec:preprocess_indicators_selection}

To prepare the input data for the machine learning model, the average value of each trust indicator is computed for each task iteration, per participant. 
Given that human preferences are represented by binary labels (-1 for the previously preferred trajectory, 1 for the current one), we compute differences in averaged trust indicator values between the two trajectories. These differences capture the variation in average feature values between the current and previously preferred parameter sets.

A \textit{Pearson correlation} analysis is performed to identify potential redundancies among the trust indicators. The Pearson correlation coefficient $c$ between two variables $A$ and $B$ is calculated as:

\vspace{-0.2cm}
\begin{equation}
c = \frac{\sum_{h=1}^{o} (A_h - \bar{A})(B_h - \bar{B})}{\sqrt{\sum_{h=1}^{o} (A_h - \bar{A})^2} \sqrt{\sum_{h=1}^{o} (B_h - \bar{B})^2}}
\label{pearson}
\end{equation}

\noindent where $A$ and $B$ are the two trust indicators being compared, $A_h$ and $B_h$ represent the $h$-th observation of $A$ and $B$, $\bar{A}$ and $\bar{B}$ are their respective mean values, and $o$ is the total number of observations. 
The \textit{tree-based feature importance} method evaluates the contribution of trust indicators to model predictions. \textit{XGBoost} is selected for its robustness against overfitting and its boosting algorithm, which captures complex feature patterns. Feature importance is measured using the \textit{gain} metric, reflecting each feature’s average performance improvement during training. A \textit{5\%} threshold of the highest importance score filters for the most influential features in further analysis.

\begin{table*}[t]
    \vspace{0.13cm} 
    \caption{\small Hyperparameter Tuning Summary for Each Model.}
    \label{tab:hyperparameter_summary}
    \small
    \centering
    \setlength{\tabcolsep}{2pt} 
    \begin{tabular}{ccccc}
    \toprule
    \textbf{Model} & \textbf{Hyperparameter} & \textbf{Description} & \textbf{Option Values} & \textbf{Optimized Value} \\
    \midrule
    \textit{Random Forest} & max\_depth & Max depth of the tree & [None, 10, 20] & None \\
     & min\_samples\_leaf & Min samples at a leaf node & [1, 2, 4] & 1 \\
     & min\_samples\_split & Min samples to split a node & [2, 5, 10] & 2 \\
     & n\_estimators & Number of trees & [50, 100, 150] & 150 \\
    \midrule
    \textit{KNN Classifier} & n\_neighbors & Number of neighbors & [3, 5, 7, 10] & 3 \\
     & weights & Weight function & ['uniform', 'distance'] & distance \\
     & algorithm & Nearest neighbors algorithm & ['auto', 'ball\_tree', 'kd\_tree', 'brute'] & auto \\
    \midrule
    \textit{SVM} & C & Regularization & [0.1, 1, 10, 100] & 100 \\
     & kernel & Kernel type & ['linear', 'rbf', 'poly'] & rbf \\
     & gamma & Kernel coefficient & ['scale', 'auto'] & scale \\
    \midrule
    \textit{Voting Classifier} & voting method & Prediction combination method & ['hard', 'soft'] & soft \\
     & base models & Included models & [Random Forest, KNN, SVM] & Optimized configurations \\
    \bottomrule
    \end{tabular}
\end{table*}

The final preprocessing step applies data augmentation to generate synthetic samples, enhancing dataset robustness. Due to the limited size of the original dataset, this was essential to mitigate overfitting and improve generalization. In low-data regimes, models often mistake noise for signal, resulting in unreliable outcomes. To address this, we employed \textit{probabilistic sampling}, drawing synthetic data independently for each feature from \textit{univariate Gaussian distributions} fitted to the global empirical mean and variance. This preserved the dataset’s statistical structure while increasing sample density, yielding a more stable and generalizable learning process.

\subsection{Machine Learning Classification of Human Preferences}
\label{sec:machine_learning}

After identifying the relevant behavioral indicators for trust modeling, the next step is to evaluate whether these indicators can classify participants' expressed trust preferences for different robot behaviors. 
To classify human preferences for robot trajectories, we employed \textit{Random Forest}, \textit{K-Nearest Neighbors} (KNN), and \textit{Support Vector Machine} (SVM). \textit{Random Forest} reduces overfitting and handles noisy data through ensemble learning. \textit{KNN} captures local patterns by classifying based on neighbor proximity. \textit{SVM} excels in high-dimensional spaces, providing clear decision boundaries and strong generalization to distinguish subtle trust differences.
Additionally, a \textit{Voting Classifier} was implemented to combine predictions from these models, leveraging their strengths to enhance reliability, mitigate overfitting, and improve classification performance.

The dataset was split into training and testing sets, with \textit{80\%} allocated to \textit{training} and \textit{20\%} to \textit{testing}. \textit{Stratification} was applied to maintain consistent class distribution across both sets.  
With reference to hyperparameter optimization, the \textit{GridSearchCV} technique was employed to fine-tune the model's parameters. With reference to~\cite{vabalas2019machine}, \textit{Nested cross-validation} was used to assess model performance and minimize the risk of overfitting. Both the outer and inner cross-validation processes utilized \textit{Stratified K-Fold Cross-Validation} with \textit{5 folds}, preserving the class distribution across each fold. This approach is particularly advantageous for small datasets as it provides a more reliable estimate of model performance. In the inner cross-validation, \textit{GridSearchCV} explored a specified range of hyperparameters, optimizing the classifier based on accuracy. 
Meanwhile, the outer cross-validation assessed the model’s generalization to unseen data, yielding nested accuracy scores that reflect its reliability. This comprehensive approach leverages all available data while preserving the target class distribution. The tested hyperparameter ranges and the final selections for each model are summarized in Table~\ref{tab:hyperparameter_summary}.

\subsection{Model Explainability to Analyze the Influence of Behavioral Trust Indicators}
\label{sec:definition_trust_score_personalized}

To gain insights into the influence of behavioral trust indicators, we employed a model explainability technique to quantify the contribution and impact of each feature on the model's predictions. Specifically, we utilized SHAP values, which enhance transparency in the model’s decision-making process by identifying the most influential trust indicators and determining their positive or negative contributions to the predictions. 
As described in Section~\ref{sec:preprocess_indicators_selection}, the input to the model consists of differences in the averaged trust indicator values between two proposed trajectories. A positive value for an input feature indicates an increase in the trust indicator with the new proposed trajectory, while a negative value indicates a decrease. Similarly, a positive SHAP value associated with a model prediction signifies a contribution to the positive class (indicating a preference for the new trajectory), whereas a negative SHAP value reflects a contribution to the negative class (indicating a preference for the previously considered best trajectory). 
This approach allows us to evaluate the relationship between variations in trust indicators and expressed trust preferences, determining whether an increase (or decrease) in the indicator value with the proposed trajectory fosters trust (if the new trajectory is preferred) or diminishes trust (if the previous trajectory is preferred), and, consequently, whether higher or lower indicator values are more desirable.

SHAP values define weights for the behavioral trust indicators in Section~\ref{sec:human_indicators}, enabling a \textit{continuous trust score} to support online adaptation of robot behavior to foster trust. Since trust is subjective, operators may prefer different trust-enhancing behaviors, reflected in their behavioral indicators. Personalizing the \textit{continuous trust score} with SHAP values allows quantifying the contribution of each indicator for individual users.

\section{EXPERIMENTS}
\label{sec:experiments}

The following section outlines the task and experimental procedure used to evaluate the proposed framework, and describes the sensor data acquisition setup for calculating the trust indicators defined in Section~\ref{sec:indicators_definition}.\\

\subsection{Task Description and PBO Parameters}
\label{sec:task}

\begin{figure}[t]
  \vspace{0.2cm} 
    \centering
    \includegraphics[width=.56\linewidth]{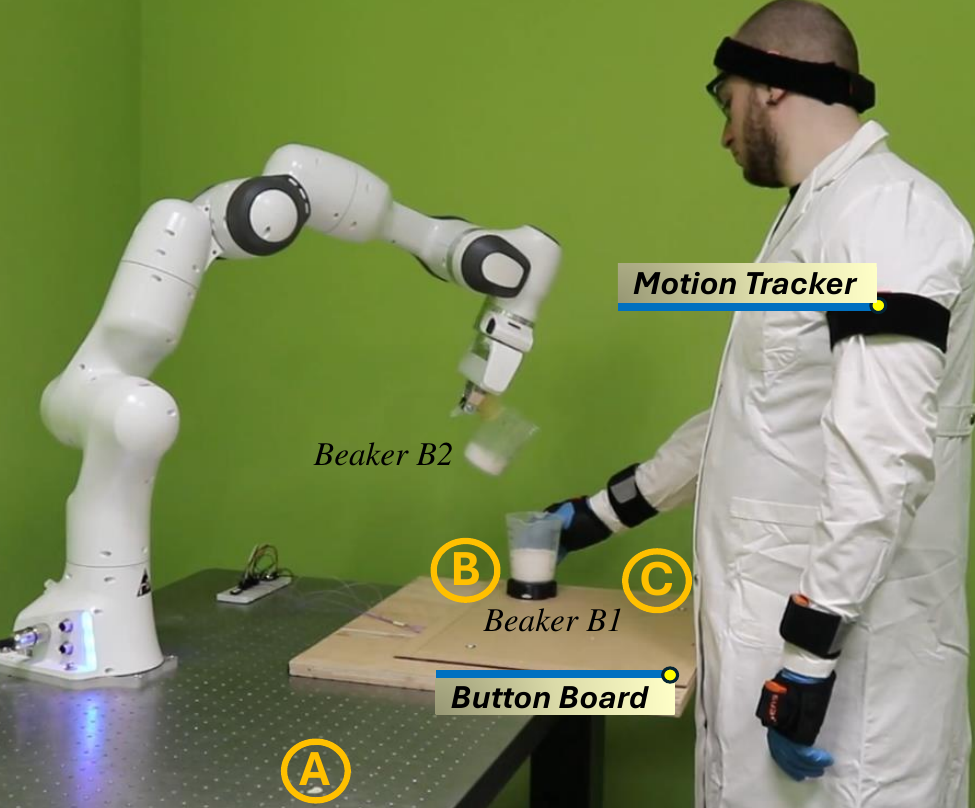}
    \caption{The experimental scenario with the robot assisting the human operator in mixing chemicals.
    \vspace{-0.3cm}}
    \label{scenario_setup}
\end{figure}

The proposed scenario featured a cyclic HRC task set in a chemical industrial environment, as illustrated in Fig.~\ref{scenario_setup}. In this setup, the robot manipulator assisted the human operator in the process of mixing chemicals. The task began with the participant placing a beaker labeled $B_1$, containing a chemical, at location (\textbf{C}). Simultaneously, the robot retrieved another beaker, $B_2$, located at (\textbf{A}) and pre-filled with a chemical. To enhance realism, participants were told that the chemicals were hazardous, though coarse salt was used as a safe substitute. Upon hearing a cue, the participant moved beaker $B_1$ from (\textbf{C}) to (\textbf{B}) and held it steady, while the robot transported beaker $B_2$ to (\textbf{B}) and poured its contents into beaker $B_1$. Once pouring was complete, the robot returned beaker $B_2$ to (\textbf{A}), and the participant returned beaker $B_1$ to (\textbf{C}). 

Regarding the interaction parameters described in Section~\ref{sec:data_collection}, the \textit{total execution time} $\tau$, representing the time taken for the robot to move beaker $B2$ to the pouring position \textbf{B}, was optimized by the PBO within the range $[5,10]$s. The \textit{separation distance} $d$ varied within $[0.52,0.65]$m, while the \textit{maximum height} $h$ corresponded to the pouring height, ranging from $[0.26, 0.40]$m. 
The maximum separation distance \( d_{\text{max}} \) was estimated empirically from preliminary trials based on the maximum observed head displacement during task execution. A similar procedure was used to determine \(\text{v}_{\text{min/max},\text{hand}}\) and \(\text{v}_{\text{min/max},\text{EE}}\), based on the observed velocity ranges of the human hand and the robot end-effector, respectively.

\subsection{Experimental Setup}
\label{sec:setup}

The robot manipulator used in this study was the \textit{Franka Emika Panda}, equipped with a two-fingered parallel gripper and controlled at 1 kHz via the \textit{Robot Operating System} (ROS). 

The \textit{Xsens MVN Awinda} motion tracking system was utilized to record the initial head position \(\mathbf{p}_{\text{init}}\) following calibration and to monitor head and body movements throughout the task. 
A button located at (\textbf{C}) detected the moment when beaker \(B_1\) was raised (indicated by the button being unpressed), marking the initiation time of human motion (\(t_H\)).

\subsection{Experimental Protocol}
\label{sec:protocol}

The study included $14$ healthy participants ($8$ males and $6$ females, aged $29.21 \pm 4.90$ years), recruited from the student body and research staff at the Istituto Italiano di Tecnologia\footnote{Experiments were carried out in accordance with the Helsinki Declaration, with the protocol receiving approval from the ASL3 Genovese ethics committee (Protocol IIT\_HRII\_ERGOLEAN 156/2020).}.

Participants completed the task $16$ times, providing feedback after each cycle for a total of 15 evaluations. Initially, $2$ cycles with different 
parameter sets were conducted, and participants indicated their preference (\textit{feedback}) between them. Subsequently, at the end of each cycle, participants compared the current parameter set proposed by the PBO to their preferred one so far. If needed, they could repeat the robot trajectory with the previous preferred set.

\section{EXPERIMENTAL RESULTS}
\label{sec:results}

This section presents key findings from evaluating the machine learning model that categorizes human preferences for trust-enhancing robot behaviors in industrial HRC. The evaluation assesses predictive accuracy and reliability using human- and robot-related trust indicators, and examines how these behavioral indicators influence predictions and shape trust levels.

\subsection{Performance Analysis of Trust Classification Models}
\label{sec:machine_learning_evaluation}

Before evaluating the model, the dataset underwent comprehensive pre-processing (detailed in Section~\ref{sec:preprocess_indicators_selection}) to create a compact and meaningful dataset. The \textit{Pearson correlation} analysis revealed a correlation coefficient of \( r = 0.65 \) between the two trust indicators, \textit{Human Attention to Task} and \textit{Human Attention to End-Effector}, indicating redundancy in the model input data. For effective feature selection, a \textit{tree-based feature importance} analysis using \textit{XGBoost} was conducted to assess the contribution of each indicator. While all features were initially considered valid, \textit{Human Attention to End-Effector} was excluded due to its redundancy, as highlighted by the correlation analysis, and its lower importance in the feature importance evaluation. To enhance robustness, \textit{data augmentation} was performed on the initial dataset of $210$ samples. A total of $630$ synthetic samples were generated (three times the original size), striking a balance between increasing sample size and preserving inherent data patterns. \textit{Nearest Neighbor Labeling} assigned labels to synthetic samples based on the majority vote of the $5$ nearest original data points. This process increased the dataset size to $840$ samples. To ensure class balance and avoid bias, \textit{Synthetic Minority Over-sampling Technique} was applied to generate additional samples for the minority class, achieving a balanced class distribution in the augmented dataset.

\begin{figure}[t]
\begin{subfigure}{0.44\linewidth}
\centering
  \includegraphics[width=\linewidth]{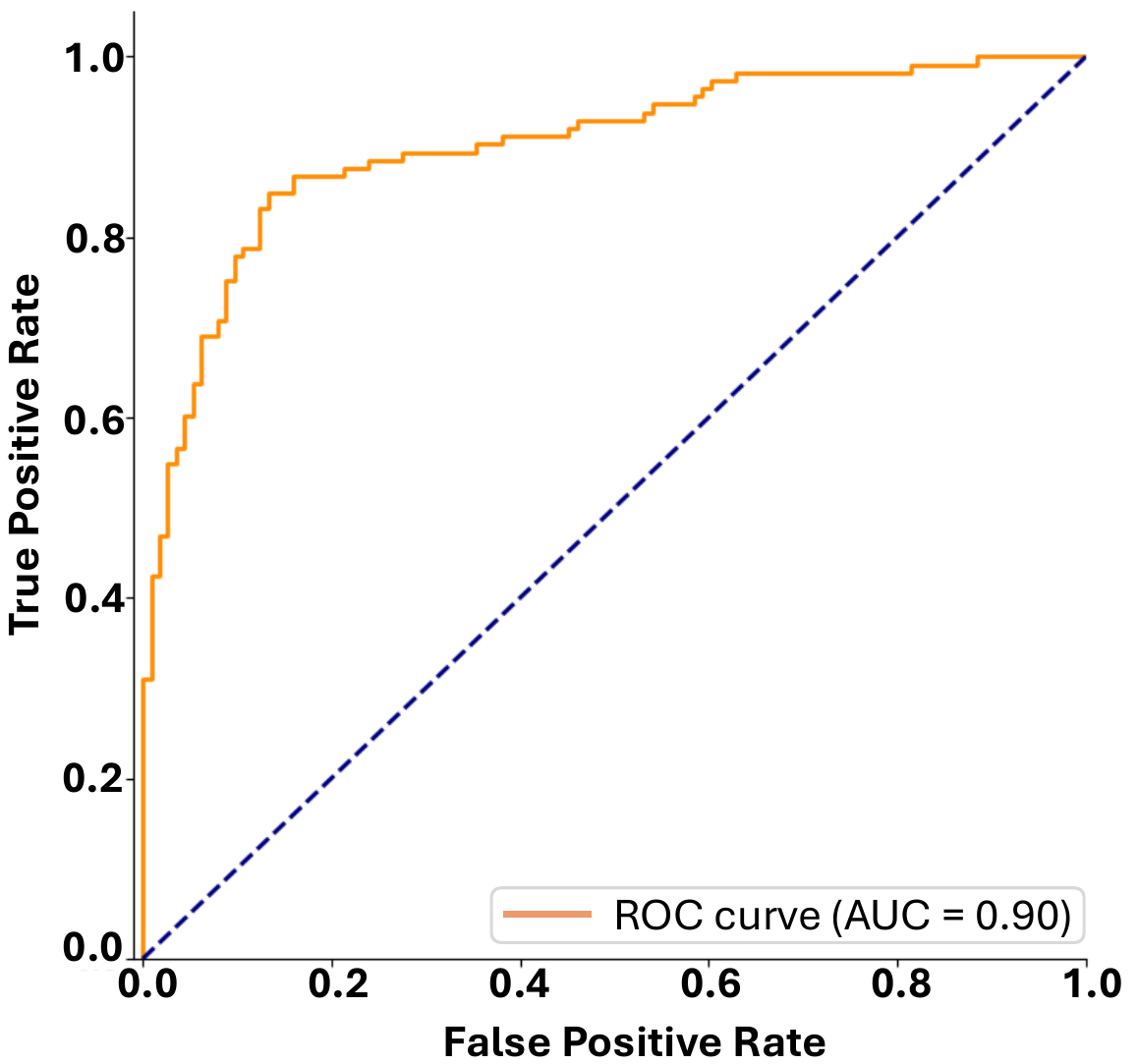}
  \caption{
  ROC Curve with an AUC value of 0.90.} 
  \label{AUC_ROC}
\end{subfigure}
\hfill
\begin{subfigure}{0.52\linewidth}
\centering
  \includegraphics[width=\linewidth]{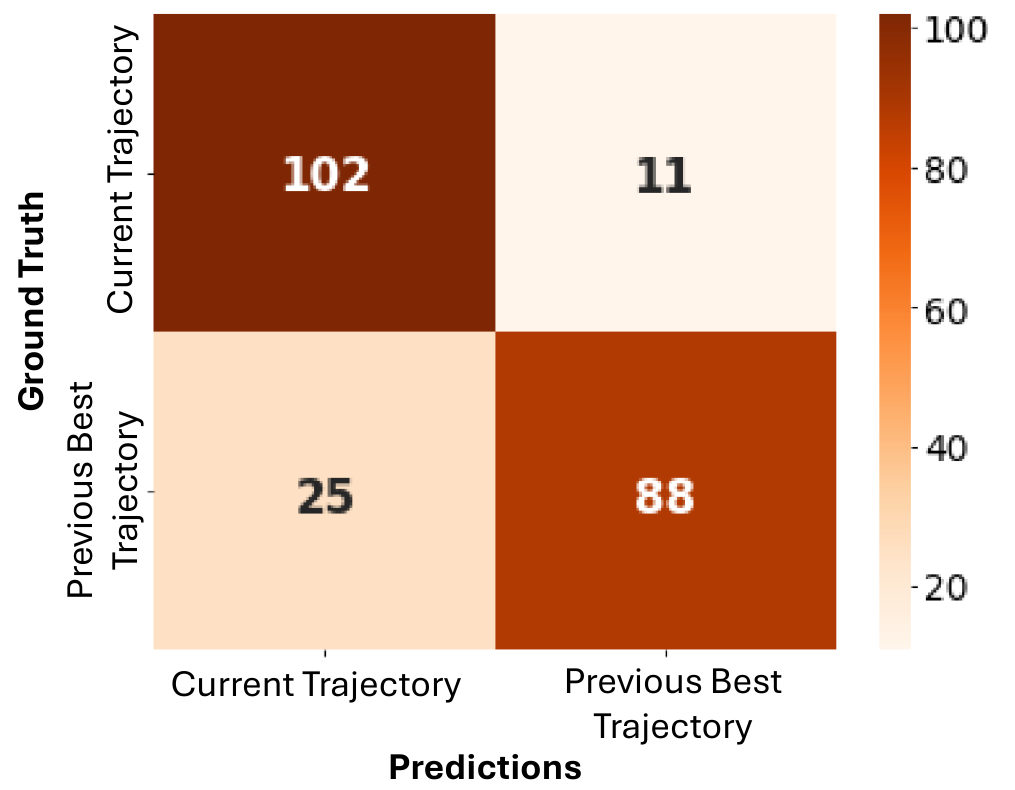}
  \caption{
  Confusion Matrix.}
  \label{confusion_matrix}
\end{subfigure}
\caption{\small 
Evaluation of the Voting Classifier.
\vspace{-0.3cm}}
\label{analysis_results}
\end{figure}

Table~\ref{performance_indicators} summarizes the performance of all tested models. The \textit{Voting Classifier} was the most robust, achieving an accuracy of 84.07\%, surpassing individual models in both classification accuracy and stability. Its performance was further validated with the \textit{Receiver Operating Characteristic} (ROC) curve (Fig.\ref{AUC_ROC}), which illustrates the true positive and false positive rates, and an \textit{Area Under the Curve} (AUC) of 0.90, confirming strong class distinction. The confusion matrix (Fig.\ref{confusion_matrix}) shows true/false positives and negatives. Additionally, the model achieved a \textit{precision} of 0.85, \textit{recall} of 0.84, and an \textit{F1 score} of 0.84, indicating a balanced performance with reliable positive identification and a good trade-off between false positives and false negatives.

\subsection{Model Explainability with SHAP Analysis}
\label{sec:shap_analysis}

\begin{figure}[t]
    \vspace{0.1cm}
    \centering
    \includegraphics[width=.90\linewidth]{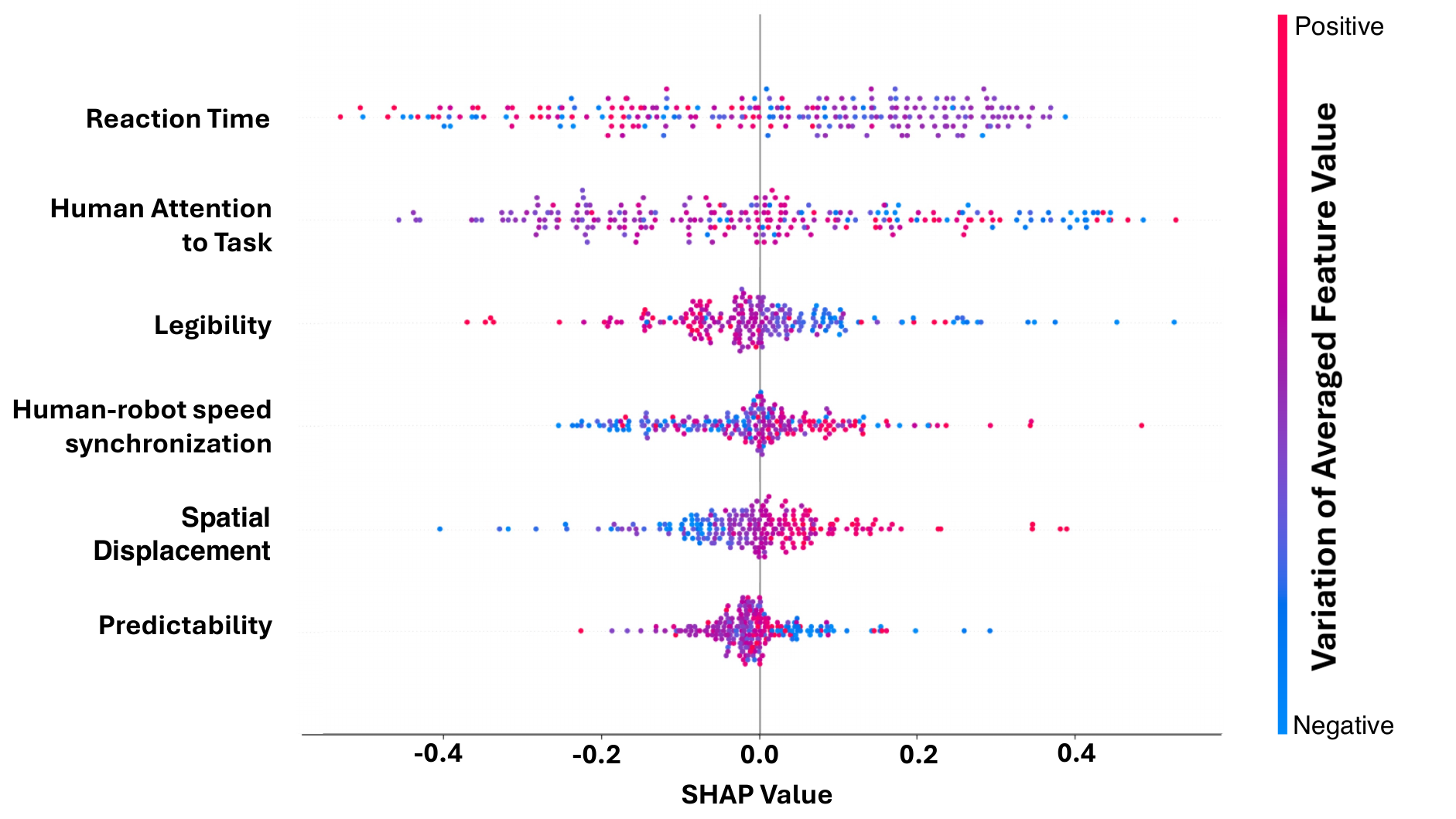}
    \caption{\small SHAP plot showing feature contributions to model predictions. 
    }
    \label{shap_overall}
\end{figure}

Fig.~\ref{shap_overall} illustrates the explainability of the model using SHAP values, which reveal the contributions of individual features to the \textit{Voting Classifier}’s decision-making process. Specifically, the figure highlights the influence of behavioral indicators on human trust preferences in robot trajectories. Each dot represents a prediction, with its color indicating the feature value. Positive SHAP values indicate predicted preference for the new trajectory, while negative values reflect preference for the previous best trajectory. Features are ranked by impact, with \textit{Reaction Time} and \textit{Human Attention to Task} being the most influential, with average absolute SHAP values of $0.197$ and $0.183$, respectively. 

\begin{table}[t]
    \vspace{0.2cm}
    \caption{\small Machine learning models and performance indicators.
    }
    \label{performance_indicators}
    \small
    \centering
    \setlength{\tabcolsep}{3pt}
    \begin{tabular}{cccccc} 
    \toprule[1.5 pt]
    \textbf{Model} & \textbf{Accuracy} & \textbf{AUC} & \textbf{Precision} & \textbf{Recall} & \textbf{F1-score}\\
    \midrule
    Random Forest & 80.09\% & 0.87 & 0.81 & 0.80 & 0.80 \\
    KNN Classifier & 80.97\% & 0.87 & 0.83 & 0.81 & 0.81 \\
    SVM & 80.02\% & 0.84 & 0.80 & 0.80 & 0.80 \\
    Voting Classifier & 84.07\% & 0.90 & 0.85 & 0.84 & 0.84 \\
    \bottomrule[1.5 pt]
    \end{tabular}
\end{table}

Notably, clusters of consistently colored points appear for each feature, suggesting that similar feature values consistently guided predictions in the same direction, which highlights the model's robustness in leveraging these indicators. 
Specifically, notable variations in \textit{Reaction Time}—including decreases (represented by light blue dots) and substantial increases (depicted by light red dots)—are associated with negative SHAP values, indicating a lower level of trust in the newly proposed trajectory. In contrast, significant reductions in \textit{Attention to the Task} (represented by light blue dots) typically correspond with a trust-driven preference for the new trajectory (associated with positive SHAP values), potentially signaling confidence in the robot's motion. 
For other indicators, the trends are more consistent and monotonic. Higher values of \textit{Human-Robot Speed Synchronization} correlate with greater trust, suggesting that human operators prefer new trajectories that facilitate synchronization. For \textit{Spatial Displacement}, increased movement by the human is associated with higher trust, reflecting a preference for the new trajectory, whereas reduced movement or freezing corresponds to diminished trust. Contrary to expectations, for \textit{Legibility} and \textit{Predictability}, significant increases in feature values (such as larger curvature radii or less jerky motions) do not consistently lead to an enhancement in trust, possibly because human operators may prioritize rapid, direct movements over smoother trajectories, which could be perceived as less efficient or slower. 
Note that similar, though expectedly less distinct, feature trends were identified in the SHAP analysis conducted without data augmentation.

\begin{figure}[t]
\vspace{-0.2cm}
\begin{subfigure}{\linewidth}
\centering
  \includegraphics[width=0.80\linewidth]{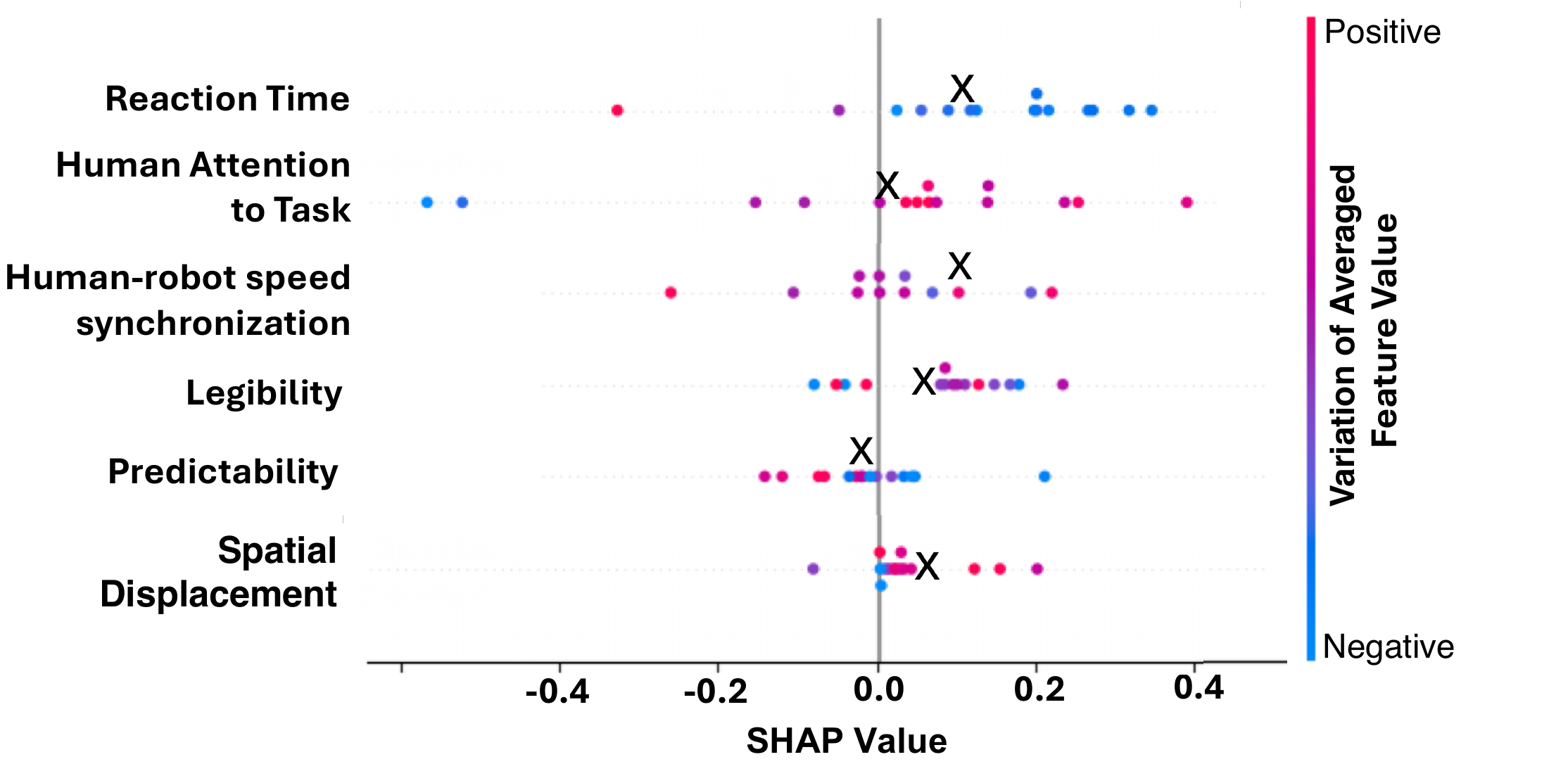}
  \vspace{-0.2cm} 
  \caption{Impact of trust indicators for participant 6.} 
  \vspace{0.3cm} 
  \label{shap_participant_6}
\end{subfigure}
\vspace{0.3cm} 
\begin{subfigure}{\linewidth}
\centering
  \includegraphics[width=0.81\linewidth]{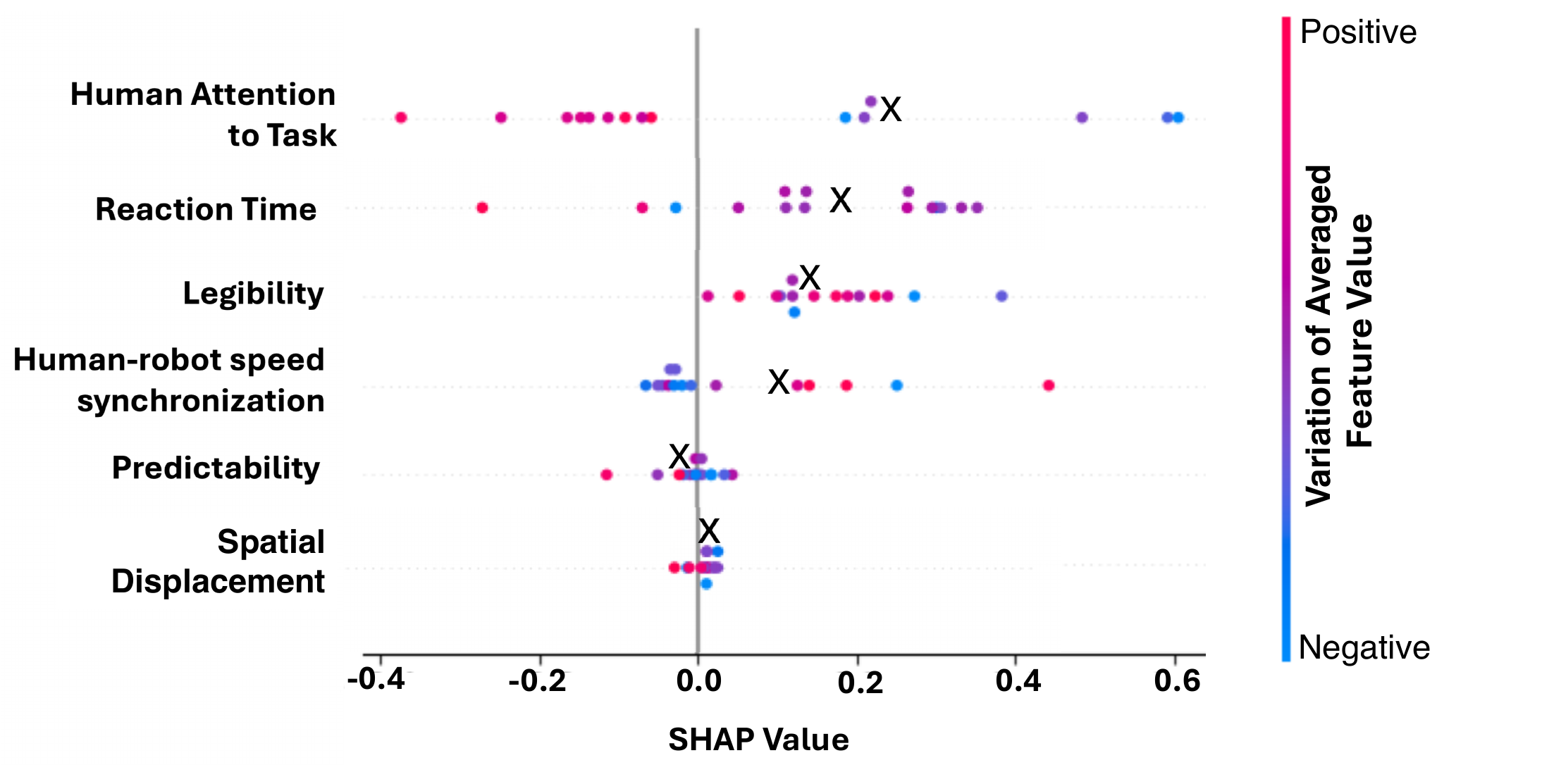}
  \vspace{-0.2cm} 
  \caption{Impact of trust indicators for participant 14.}
  \label{shap_participant_9}
\end{subfigure}
\caption{\small Importance of trust-related behavioral indicators for two participants, with X representing the mean SHAP values.
\vspace{-0.4cm}} 
\label{shap_participants}
\end{figure}
 
Given the subjectivity inherent in trust and body language indicators, the analysis was also conducted on a per-participant basis. 
The importance of trust indicators for two different participants is shown in Fig.~\ref{shap_participants}. 
For Participant 6, the most influential trust indicators were \textit{Reaction Time} (average absolute SHAP = $0.186$), where a decrease in value indicated higher trust in the robot, and \textit{Human Attention to Task} (average absolute SHAP = $0.185$), where an increase in value signaled a preference for the new trajectory. Interestingly, this deviated from the general trend observed across all participants in Fig.~\ref{shap_overall}. 
Conversely, for Participant 14, the most influential feature was the variation in averaged \textit{Human Attention to Task} (average absolute SHAP = $0.247$), where lower attention was correlated with higher trust. The second most influential feature was \textit{Reaction Time} (average absolute SHAP = $0.202$), which similarly indicated higher trust at lower values.

\section{DISCUSSION and CONCLUSIONS}
\label{sec:discussion}

This paper presented a data-driven method for estimating trust based on human- and robot-related behavioral indicators. Using PBO, we gathered explicit user feedback on preferences between pairs of interaction parameters that determined the robot’s trajectory in a collaborative chemical task. These preferences served as labels for a machine learning model trained to predict trust preferences based solely on the proposed indicators.

Experimental results showed that machine learning models using human behavioral indicators can effectively predict human trust preferences in HRC (\textbf{RQ\ref{RQ1}}). Among the models tested, the \textit{Voting Classifier} outperformed individual models such as Random Forest, KNN, and SVM, achieving an accuracy of \textit{84.07\%} and an $AUC$ of \textit{0.90} in selecting the preferred robot trajectory.  
This model could enable the automatic optimization of robot interaction parameters—such as human-robot separation distance, robot velocity, and vertical distance from the head—based on body language analysis, eliminating the need for direct user feedback.

The model explainability analysis revealed general trends in the proposed behavioral indicators and their relationship to trust-related preferences in robot trajectories (\textbf{RQ\ref{RQ2}}). This analysis identified consistent variations (increases or decreases) in behavioral indicators that signify an increase in trust across participants, revealing high or low values of specific indicators are preferable for promoting trust. 
Additionally, the subject-specific analysis of SHAP values highlighted the varying importance of trust-related features across participants, emphasizing the need for personalized trust models. 
This work lays the foundation for a continuous, personalized trust model that computes normalized trust indicators online and combines them into a weighted sum, using absolute average SHAP values and trend-based signs. The resulting online trust estimate informs the robot, enabling it to evaluate whether its adaptations and optimizations positively influence human trust.

Despite its promising contributions, the study has limitations. The framework was tested in a controlled pouring task within a chemical industry scenario, with researchers overseeing the process. A natural next step is to 
validate this approach across a broader range of tasks, particularly since the methodology was designed to be applicable to various HRC scenarios involving close-proximity interaction and robot trajectory planning. Additionally, we plan to deploy and validate the online model for trust assessment and trust-driven robot behavior adaptation, with the goal of fostering mutual understanding and trustworthy collaboration. 
Further research will also examine the integration of advanced learning models, such as deep learning, to enhance system performance and adaptability.

\balance

\vspace{-0.2cm}
\bibliographystyle{IEEEtran}
\bibliography{biblio}

@article{campagna2024promoting,
  title={Promoting Trust in Industrial Human-Robot Collaboration through Preference-Based Optimization},
  author={Campagna, Giulio and Lagomarsino, Marta and Lorenzini, Marta and Chrysostomou, Dimitrios and Rehm, Matthias and Ajoudani, Arash},
  journal={IEEE Robotics and Automation Letters},
  year={2024},
  publisher={IEEE}
}

@inproceedings{weidenbacher2007comprehensive,
  title={A comprehensive head pose and gaze database},
  author={Weidenbacher, Ulrich and Layher, Georg and Strauss, P-M and Neumann, Heiko},
  booktitle={2007 3rd IET International Conference on Intelligent Environments},
  pages={455--458},
  year={2007},
  organization={IET}
}

@article{lagomarsino2022pick,
  title={Pick the right co-worker: Online assessment of cognitive ergonomics in human--robot collaborative assembly},
  author={Lagomarsino, Marta and Lorenzini, Marta and Balatti, Pietro and De Momi, Elena and Ajoudani, Arash},
  journal={IEEE Transactions on Cognitive and Developmental Systems},
  volume={15},
  number={4},
  pages={1928--1937},
  year={2022},
  publisher={IEEE}
}

@inproceedings{campagna2023analysis,
  title={Analysis of proximity and risk for trust evaluation in human-robot collaboration},
  author={Campagna, Giulio and Rehm, Matthias},
  booktitle={2023 32nd IEEE International Conference on Robot and Human Interactive Communication (RO-MAN)},
  pages={2191--2196},
  year={2023},
  organization={IEEE}
}

@inproceedings{campagna2024data,
  title={A Data-Driven Approach Utilizing Body Motion Data for Trust Evaluation in Industrial Human-Robot Collaboration},
  author={Campagna, Giulio and Dadgostar, Mahed and Chrysostomou, Dimitrios and Rehm, Matthias},
  booktitle={33rd IEEE International Conference on Robot and Human Interactive Communication, IEEE RO-MAN 2024},
  year={2024},
  organization={IEEE}
}

@inproceedings{bartkowski2023sync,
  title={In Sync: Exploring Synchronization to Increase Trust Between Humans and Non-humanoid Robots},
  author={Bartkowski, Wieslaw and Nowak, Andrzej and Czajkowski, Filip Ignacy and Schmidt, Albrecht and M{\"u}ller, Florian},
  booktitle={Proceedings of the 2023 CHI Conference on Human Factors in Computing Systems},
  pages={1--14},
  year={2023}
}

@article{khan2006inferring,
  title={Inferring online and offline processing of visual feedback in target-directed movements from kinematic data},
  author={Khan, Michael A and Franks, Ian M and Elliott, Digby and Lawrence, Gavin P and Chua, Romeo and Bernier, Pierre-Michel and Hansen, Steve and Weeks, Daniel J},
  journal={Neuroscience \& Biobehavioral Reviews},
  volume={30},
  number={8},
  pages={1106--1121},
  year={2006},
  publisher={Elsevier}
}

@inproceedings{lagomarsino2022robot,
  title={Robot trajectory adaptation to optimise the trade-off between human cognitive ergonomics and workplace productivity in collaborative tasks},
  author={Lagomarsino, Marta and Lorenzini, Marta and De Momi, Elena and Ajoudani, Arash},
  booktitle={2022 IEEE/RSJ International Conference on Intelligent Robots and Systems (IROS)},
  pages={663--669},
  year={2022},
  organization={IEEE}
}

@inproceedings{dragan2013legibility,
  title={Legibility and predictability of robot motion},
  author={Dragan, Anca D and Lee, Kenton CT and Srinivasa, Siddhartha S},
  booktitle={2013 8th ACM/IEEE International Conference on Human-Robot Interaction (HRI)},
  pages={301--308},
  year={2013},
  organization={IEEE}
}

@article{bemporad2021global,
  title={Global optimization based on active preference learning with radial basis functions},
  author={Bemporad, Alberto and Piga, Dario},
  journal={Machine Learning},
  volume={110},
  pages={417--448},
  year={2021},
  publisher={Springer}
}

@article{gervasi2023experimental,
  title={An experimental focus on learning effect and interaction quality in human--robot collaboration},
  author={Gervasi, Riccardo and Mastrogiacomo, Luca and Franceschini, Fiorenzo},
  journal={Production Engineering},
  volume={17},
  number={3},
  pages={355--380},
  year={2023},
  publisher={Springer}
}

@article{lee2004trust,
  title={Trust in automation: Designing for appropriate reliance},
  author={Lee, John D and See, Katrina A},
  journal={Human factors},
  volume={46},
  number={1},
  pages={50--80},
  year={2004},
  publisher={SAGE Publications Sage UK: London, England}
}

@article{de2020towards,
  title={Towards a theory of longitudinal trust calibration in human--robot teams},
  author={De Visser, Ewart J and Peeters, Marieke MM and Jung, Malte F and Kohn, Spencer and Shaw, Tyler H and Pak, Richard and Neerincx, Mark A},
  journal={International journal of social robotics},
  volume={12},
  number={2},
  pages={459--478},
  year={2020},
  publisher={Springer}
}

@incollection{schaefer2016measuring,
  title={Measuring trust in human robot interactions: Development of the “trust perception scale-HRI”},
  author={Schaefer, Kristin E},
  booktitle={Robust intelligence and trust in autonomous systems},
  pages={191--218},
  year={2016},
  publisher={Springer}
}

@article{campagna2025systematic,
  title={A Systematic Review of Trust Assessments in Human--Robot Interaction},
  author={Campagna, Giulio and Rehm, Matthias},
  journal={ACM Transactions on Human-Robot Interaction},
  volume={14},
  number={2},
  pages={1--35},
  year={2025},
  publisher={ACM New York, NY}
}

@inproceedings{xu2015optimo,
  title={Optimo: Online probabilistic trust inference model for asymmetric human-robot collaborations},
  author={Xu, Anqi and Dudek, Gregory},
  booktitle={Proceedings of the tenth annual ACM/IEEE international conference on human-robot interaction},
  pages={221--228},
  year={2015}
}

@article{shayesteh2022workers,
  title={Workers’ trust in collaborative construction robots: EEG-based trust recognition in an immersive environment},
  author={Shayesteh, Shayan and Ojha, Amit and Jebelli, Houtan},
  journal={Automation and robotics in the architecture, engineering, and construction industry},
  pages={201--215},
  year={2022},
  publisher={Springer}
}

@inproceedings{campagna2024analysis,
  title={Analysis of Facial Features for Trust Evaluation in Industrial Human-Robot Collaboration},
  author={Campagna, Giulio and Chrysostomou, Dimitrios and Rehm, Matthias},
  booktitle={2024 IEEE International Conference on Advanced Robotics and Its Social Impacts (ARSO)},
  pages={1--6},
  year={2024},
  organization={IEEE}
}

@article{lagomarsino2023maximising,
  title={Maximising Coefficiency of Human-Robot handovers through reinforcement learning},
  author={Lagomarsino, Marta and Lorenzini, Marta and Constable, Merryn Dale and De Momi, Elena and Becchio, Cristina and Ajoudani, Arash},
  journal={IEEE Robotics and Automation Letters},
  volume={8},
  number={8},
  pages={4378--4385},
  year={2023},
  publisher={IEEE}
}

@InProceedings{kanda2003Body,
  author    = {Takayuki Kanda and Hiroshi Ishiguro and Michita Imai and Tetsuo Ono},
  booktitle = {Proceedings of Int. Joint Conference on Artificial Intelligence},
  title     = {Body Movement Analysis of Human-Robot Interaction},
  year      = {2003},
}

@article{lagomarsino2024promind,
  author    = {Lagomarsino, Marta and Lorenzini, Marta and Ajoudani, Arash},
  title     = {{PRO-MIND: Proximity and Reactivity Optimisation of robot Motion to tune safety limits, human stress, and productivity in INDustrial setting}},
  journal={IEEE Transactions on Robotics},
  year={2025},
  volume={41},
  number={},
  pages={2067-2085},
}

@InBook{holmqvist2011eye,
  pages     = {187},
  title     = {Eye Tracking : A Comprehensive Guide to Methods and Measures},
  publisher = {Oxford University Press},
  year      = {2011},
  author    = {Kenneth Holmqvist and Marcus Nystr{\"o}m and Richard Andersson and Richard Dewhurst and Jarodzka Halszka and {van de Weijer}, Joost},
  isbn      = {9780199697083},
}

@InProceedings{kuehnlenz2016reduction,
  author    = {Kühnlenz, Barbara and Kühnlenz, Kolja},
  title     = {Reduction of Heart Rate by Robot Trajectory Profiles in Cooperative HRI},
  booktitle = {Proceedings of International Symposium on Robotics (ISR)},
  publisher = {{IEEE}},
  year      = {2016},
  pages     = {400--406},
  isbn      = {978-3-8007-4231-8},
}

@Article{flash1985coordination,
  author    = {T Flash and N Hogan},
  title     = {The coordination of arm movements: an experimentally confirmed mathematical model},
  journal   = {The Journal of Neuroscience},
  publisher = {Society for Neuroscience},
  year      = {1985},
  pages     = {1688--1703},
  volume    = {5},
  number    = {7},
  doi       = {10.1523/jneurosci.05-07-01688.1985},
}

@incollection{sweller2011cognitive,
  title={Cognitive load theory},
  author={Sweller, John},
  booktitle={Psychology of learning and motivation},
  volume={55},
  pages={37--76},
  year={2011},
  publisher={Elsevier}
}

@article{hancock2011meta,
  title={A meta-analysis of factors affecting trust in human-robot interaction},
  author={Hancock, Peter A and Billings, Deborah R and Schaefer, Kristin E and Chen, Jessie YC and De Visser, Ewart J and Parasuraman, Raja},
  journal={Human factors},
  volume={53},
  number={5},
  pages={517--527},
  year={2011},
  publisher={Sage Publications Sage CA: Los Angeles, CA}
}

@article{vabalas2019machine,
  title={Machine learning algorithm validation with a limited sample size},
  author={Vabalas, Andrius and Gowen, Emma and Poliakoff, Ellen and Casson, Alexander J},
  journal={PloS one},
  volume={14},
  number={11},
  pages={e0224365},
  year={2019},
  publisher={Public Library of Science San Francisco, CA USA}
}

\end{document}